\documentclass[conference]{IEEEtran}
\IEEEoverridecommandlockouts  
\usepackage[
  letterpaper,
  left=0.75in,
  right=0.75in,
  top=1in,
  bottom=0.75in
]{geometry}

\makeatletter
\newcount\mypagecount
\AddToHook{shipout/after}{%
  \global\advance\mypagecount by 1\relax
  \ifnum\mypagecount=1\relax
    \global\advance\topmargin by -0.25in\relax
    \global\advance\textheight by 0.25in\relax
    \global\@colht=\textheight
    \global\@colroom=\textheight
    \global\vsize=\textheight
  \fi
}
\makeatother

\usepackage{afterpage}
\usepackage{cite}
\usepackage{amsmath,amssymb,amsfonts}
\usepackage{graphicx}
\usepackage{textcomp}
\usepackage{xcolor}
\usepackage{booktabs}
\usepackage{bbm}
\usepackage{multirow}
\usepackage{url}
\usepackage[caption=false,font=footnotesize]{subfig}
\usepackage{siunitx}
\usepackage[hidelinks]{hyperref}

\newcommand{\method}{OCLO}
\newcommand{\vc}{\mathbf{v}_{\mathrm{cmd}}}
\newcommand{\norm}[1]{\lVert #1 \rVert}
\newcommand{\relu}[1]{\left[#1\right]_{+}}

\usepackage[table]{xcolor}   
\usepackage{makecell}        
\usepackage{booktabs}        
\usepackage{graphicx}        

\definecolor{rowhl}{HTML}{EAF2FB}   
\definecolor{headhl}{HTML}{F2F2F2}  
\definecolor{okgreen}{HTML}{1B7F3B}
\definecolor{amber}{HTML}{C08400}
\definecolor{offgray}{HTML}{9AA0A6}
\newcommand{\cmark}{\textcolor{okgreen}{\checkmark}}
\newcommand{\pmark}{\textcolor{amber}{$\circ$}}
\newcommand{\nmark}{\textcolor{offgray}{\textendash}}

\usepackage{enumitem}
\setlist[itemize]{topsep=2pt, itemsep=3pt, parsep=0pt, partopsep=0pt}
\setlist[enumerate]{topsep=2pt, itemsep=2pt, parsep=0pt, partopsep=0pt}
\AtBeginDocument{%
  \setlength{\abovedisplayskip}{4pt plus 1pt minus 1pt}%
  \setlength{\belowdisplayskip}{4pt plus 1pt minus 1pt}%
  \setlength{\abovedisplayshortskip}{2pt plus 1pt minus 1pt}%
  \setlength{\belowdisplayshortskip}{3pt plus 1pt minus 1pt}%
}

\newcommand{\teaserfigureblock}{%
\begin{center}
\includegraphics[width=0.99\textwidth,pagebox=cropbox]{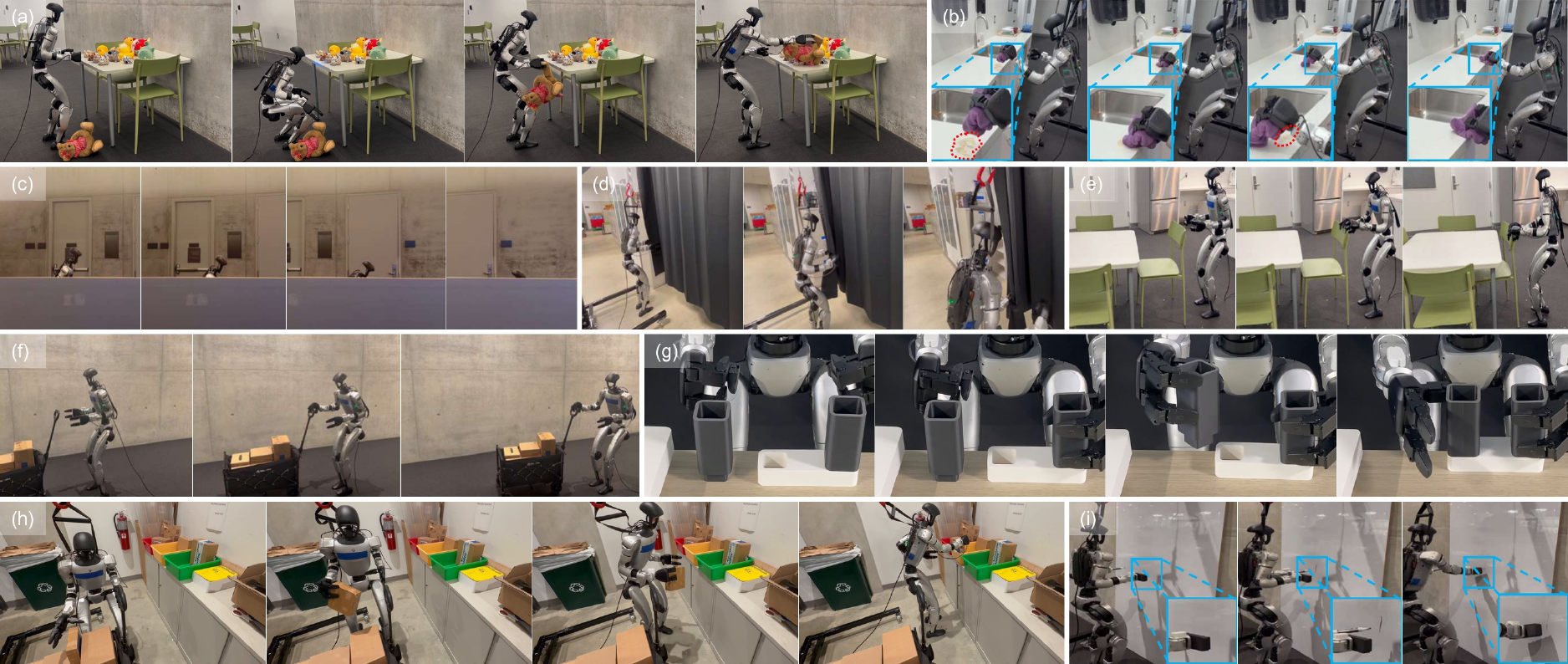}
\end{center}
\setcounter{figure}{0}%
\refstepcounter{figure}\label{fig:teaser}%
\noindent\parbox{\textwidth}{\footnotesize\textbf{Fig.~\thefigure.} \textbf{Loco-manipulation tasks performed by \method{} (\textbf{O}nline-posture \textbf{C}ompliant \textbf{LO}co-manipulation) from two hand controllers and the G1's joystick, with no posture command.} (a) Picking a doll from the ground, (b) cleaning a dirty surface, (c) escalator-like motion, (d) curtain opening, (e) chair pushing, (f) cart pulling, (g) block insertion, (h) box pick-and-place, and (i) drawing and wiping. Each panel shows representative frames from the corresponding task sequence.}%
\vspace{-3mm}%
}

\makeatletter
\let\@oldmaketitle\@maketitle
\renewcommand{\@maketitle}{\@oldmaketitle\teaserfigureblock}
\makeatother

\begin{document}

\title{\fontsize{16}{20}\selectfont\bfseries Dataset-Free Compliant Humanoid Loco-Manipulation\\ with Dynamic Online Posture}  
\author{Seungho Yeom$^{1,2}$, Zhenyu Wu$^{1*}$, Jaeyoung Huh$^{1*}$, Diego Williams$^{1*}$, Yuheng Zhi$^{1}$,\\
Soofiyan Atar$^{1}$, and Michael Yip$^{1}$%
\thanks{$^{*}$Equal contribution.}%
\thanks{$^{1}$University of California San Diego, La Jolla, CA 92093, USA.}%
\thanks{$^{2}$Yonsei University, Seoul 03722, Republic of Korea. This work was done while S. Yeom was with the University of California San Diego.}%
}
\maketitle

\begin{abstract}
Most humanoid loco-manipulation controllers require human motion data to learn whole-body coordination and posture, leaving policies reliant on external sources to provide this data. We present \method{} (Online-posture Compliant LOco-manipulation), a humanoid loco-manipulation system trained without human motion data and commanded only through two end-effector targets and a walking velocity. Because these targets do not uniquely determine whole-body posture, \method{} generates pelvis height and torso orientation online using an analytic reachability prior, further refined through policy-in-the-loop sampling with a task-agnostic cost. \method{} also learns whole-body compliance by displacing end-effector references according to measured forces through a spring-damper model, encouraging the legs, waist, and pelvis to yield to external loads. 
In simulation, using the reachability prior leads to a 77.8\% success rate in acquiring the commanded reference, a vast improvement over the 37.8\% success rate accomplished without the prior.
Further, refinement reduces end-effector orientation error across all evaluated tasks. The same posture module improves a pretrained SONIC controller on all five tasks. Without compliance training, policies tend to lose balance under disturbances rather than sacrifice tracking. On a Unitree G1, \method{} maintains balance under end-effector disturbances that cause its ablations to fail and performs nine loco-manipulation tasks, including crouched walking and picking up a box from a low surface. Project website: \href{https://oclo-humanoid.github.io/}{oclo-humanoid.github.io}
\end{abstract}

\section{Introduction}
\label{sec:intro}
A humanoid that manipulates while standing must decide not only where its hands should go, but also how the rest of the body should support them. Manipulation intent can come from many sources: an operator can stream hand poses, a script can specify geometric trajectories, or a learned policy can output end-effector targets from perception and language. These interfaces specify the hands, but not the related requirements for the rest of the body, for example, how low the pelvis should sit, how far the torso should lean, or how posture should change under contact.

Most humanoid loco-manipulation controllers resolve this ambiguity through human motion priors~\cite{omnih2o,sonic,gmt}. A human motion prior provides both posture and whole-body coordination, but requires motion data and often leaves posture tied to an external source, such as head or chest tracking~\cite{omnih2o,mobiletv} or a posture generator learned from human motion~\cite{amo}. Most compliant humanoid controllers also learn from human motion, and their posture still comes from a reference motion or an upstream root command~\cite{softmimic,gentlehumanoid,ceer}. Controllers trained without human data instead expose posture as an explicit operator command~\cite{homie,hugwbc,ulc}, and planners that do choose posture optimize a task-specific objective~\cite{r2s2,egovcp,sumo}. Without such a source, low or extended end-effector targets can become difficult to reach from a fixed nominal stance.

We present an approach to loco-manipulation that does not require obtaining posture from human motion, an external command, or a task-specific planner. To this end we separate posture generation from physical support. The control policy is responsible for maintaining balance under interaction, while the posture itself must be generated online from the end-effector targets available at deployment. This separation is important because the problem is not purely geometric: the upstream source specifies where the hands should go, but only the robot experiences the payloads and contact forces that affect balance. The generated posture must therefore make the targets reachable while allowing the controller to remain stable under interaction.

We realize this separation with \textbf{\method{}} 
(\textbf{O}nline-posture \textbf{C}ompliant \textbf{LO}co-manipulation), 
a humanoid loco-manipulation system with four main components. First, all training commands, including end-effector targets, interaction forces, and locomotion commands, are procedurally sampled. Second, OCLO realizes whole-body compliance through force-displaced EE references and a lower-body policy trained to accommodate the resulting motion: measured end-effector forces displace the commanded references through a spring-damper model, and the body is trained to yield to that displacement. Third, posture is generated online by an analytic reachability prior over pelvis height and torso orientation, and can be refined by policy-in-the-loop sampling that evaluates candidate postures using the frozen controller without any task-specific cost. Finally, we demonstrate the resulting system on a Unitree G1 using two hand controllers and the G1's joystick, with no explicit pelvis or torso command.

The two-end-effector interface is intended to separate upper-body intent from lower-body support. In this work we demonstrate the interface through teleoperation; learned upstream policies are left to future work (Sec.~\ref{sec:conclusion}).

Our experiments isolate the contribution of each component. In simulation, the analytic prior substantially improves reference acquisition compared with a fixed nominal posture and with policy-in-the-loop sampling alone, while refinement reduces end-effector orientation error across all evaluated tasks. A policy trained on end-effector targets alone learns to crouch but succeeds less often than the prior on four of five tasks. The same posture module also improves a pretrained SONIC controller~\cite{sonic}. Policies trained without randomized end-effector targets and force perturbations are substantially less robust to disturbances, particularly those applied through the arms. On a Unitree G1, \method{} withstands end-effector disturbances that its ablations cannot and performs nine loco-manipulation tasks, including crouched walking, low-object pickup, cart manipulation, and full-height board interaction, all without explicit posture commands (Fig.~\ref{fig:teaser}).

Overall, the key technical contributions of this paper are:
\begin{itemize}
    \item \textbf{Human-motion-free compliant whole-body training.}
    As in prior procedurally trained controllers~\cite{homie,hugwbc,ulc}, all training commands are sampled and no human motion data is used. We add compliance to this setting. Sampled end-effector forces displace the commanded references through a spring-damper model whose randomized parameters the policy observes, and the legs, waist, and pelvis are rewarded for tracking the displaced references, so the robot learns to yield to external loads through coordinated whole-body motion. 

    \item \textbf{Online posture generation from two end-effector targets.}
    We introduce an analytic reachability prior over pelvis height and torso orientation, followed by optional policy-in-the-loop posture refinement using the frozen controller. The prior guides the search toward postures that make both end-effectors reachable, while refinement evaluates candidate postures using a task-agnostic cost over stability, compliance, and effort, with no task reward or task-specific value function as used in prior posture planners~\cite{r2s2,egovcp}.  The same posture-generation module also improves a controller pretrained on human motion.
\end{itemize}

\section{Related Work}
\label{sec:related_work}

\method{} has four properties. It is trained without human motion data, compliance is trained into its policy, its posture is chosen online, and it performs loco-manipulation on a physical humanoid. Table~\ref{tab:positioning} positions prior systems along these
four axes. 

\begin{table}[t]
  \centering
  \caption{\textbf{Positioning against prior systems.}
\cmark\ yes, \pmark\ partial, \nmark\ no.
No human data: no motion capture, retargeted motion, or demonstrations;
Compliance: force-yielding or impedance tracking;
Posture online: robot-selected pelvis height or torso orientation;
Humanoid hardware: physical humanoid loco-manipulation.}
  \label{tab:positioning}
  \footnotesize
  \setlength{\tabcolsep}{6pt}
  \renewcommand{\arraystretch}{1.25}
  \resizebox{\linewidth}{!}{%
  \begin{tabular}{@{}lcccc@{}}
  \toprule
  \rowcolor{headhl}
  \textbf{System} &
  \makecell{\textbf{No human}\\\textbf{data}} &
  \textbf{Compliance} &
  \makecell{\textbf{Posture}\\\textbf{online}} &
  \makecell{\textbf{Humanoid}\\\textbf{hardware}} \\
  \midrule
  OmniH2O~\cite{omnih2o}                & \nmark & \nmark & \nmark & \cmark \\
  Mobile-TeleVision~\cite{mobiletv}     & \pmark & \nmark & \nmark & \cmark \\
  SONIC~\cite{sonic}                    & \nmark & \nmark & \nmark & \cmark \\
  HOMIE~\cite{homie}                    & \cmark & \nmark & \nmark & \cmark \\
  AMO~\cite{amo}                        & \pmark & \nmark & \nmark & \cmark \\
  HugWBC~\cite{hugwbc}                  & \cmark & \nmark & \nmark & \cmark \\
  ULC~\cite{ulc}                  & \cmark & \nmark & \nmark & \cmark \\
  FALCON~\cite{falcon}                  & \pmark & \pmark & \nmark & \cmark \\
  FACET~\cite{facet}                    & \cmark & \cmark & \nmark & \nmark \\
  GentleHumanoid~\cite{gentlehumanoid}  & \nmark & \cmark & \nmark & \cmark \\
  SoftMimic~\cite{softmimic}            & \nmark & \cmark & \nmark & \cmark \\
  CEER~\cite{ceer}                & \nmark & \cmark & \nmark & \cmark \\
  Unified force policy~\cite{unifiedforce} & \cmark & \cmark & \nmark & \nmark \\
  Sumo~\cite{sumo}                      & \cmark & \nmark & \cmark & \nmark \\
  R2S2~\cite{r2s2}                & \cmark & \nmark & \pmark & \cmark \\
  Ego-VCP~\cite{egovcp}           & \cmark & \nmark & \pmark & \cmark \\
  \midrule
  \rowcolor{rowhl}
  \textbf{\method{}}                    & \cmark & \cmark & \cmark & \cmark \\
  \bottomrule
  \end{tabular}%
  }
  \end{table}

\textbf{Learning from human motion.}
Most humanoid loco-manipulation controllers are trained as motion trackers from human data. OmniH2O~\cite{omnih2o}, SONIC~\cite{sonic}, GMT~\cite{gmt}, ExBody~\cite{exbody}, and ResMimic~\cite{resmimic} follow this paradigm, while Mobile-TeleVision~\cite{mobiletv} and AMO~\cite{amo} retain human-motion priors for upper-body or posture generation. HOMIE~\cite{homie}, HugWBC~\cite{hugwbc}, and ULC~\cite{ulc} avoid human data through procedural training, but posture remains an explicit command and none trains whole-body compliance. We compare against pretrained SONIC~\cite{sonic} and also apply our posture module to its controller.

\textbf{Posture as an external command.}
Existing systems typically obtain posture from an external source: the operator in HOMIE~\cite{homie} and HugWBC~\cite{hugwbc}, a hand-derived intermediate command in AMO~\cite{amo}, a root-height input in FALCON~\cite{falcon} and CEER~\cite{ceer}, or a higher-level agent in Walk These Ways~\cite{wtw}. Other unified controllers retain the stance learned beneath task-frame EE targets~\cite{umionlegs,deepwbc}. Closer to our setting, R2S2~\cite{r2s2} trains task-specific planners that select body height and torso bend over a frozen, human-motion-free skill space. In contrast, \method{} generates posture online from the EE targets alone, with no task-specific objective, and trains the whole body to yield under interaction loads.

\textbf{Compliance and sampling-based refinement.}
Our compliance formulation follows impedance control~\cite{hogan,khatib} through force-dependent reference displacement, as used in related form by FACET~\cite{facet} and GentleHumanoid~\cite{gentlehumanoid}. Prior humanoid systems incorporate force through critic-only signals~\cite{falcon}, motion-derived compliance~\cite{softmimic,gentlehumanoid}, or explicit arm control modes~\cite{hmc}; FACET~\cite{facet} and Zhi et al.~\cite{unifiedforce} train force-aware policies without human data but do not demonstrate humanoid loco-manipulation. CEER~\cite{ceer} is the closest interface to ours. Its G1 policy, distilled from a teacher trained on human motion, tracks compliant end-effector and root-pose commands with fixed impedance gains and no force observation at deployment, and its upstream source must supply the root pose, including height. Sampling-based planners can add online decisions above learned controllers without retraining~\cite{mjpc,rgbmppi}. Most closely, Sumo~\cite{sumo} searches a pretrained policy's command space, including torso posture, toward task-specific goals on Spot hardware and on a simulated G1. Ego-VCP~\cite{egovcp} runs CEM above a frozen controller on a Unitree G1, choosing end-effector positions and body height through a learned world model and a task-specific value function. \method{} instead searches only the posture interface with a task-agnostic cost, refining a closed-form prior that can also run alone.

\section{Method}
\label{sec:method}

\method{} is a loco-manipulation system whose only learned component is a lower-body policy
$\pi_\theta(a_t\mid o_t)$ that outputs Proportional-Derivative (PD) control joint targets for the twelve leg and three waist joints at
\SI{50}{\hertz}. The fourteen arm joints are controlled by differential inverse kinematics, which tracks
a compliant reference for each end-effector (EE) using the same solver in training and on hardware.

 The policy observes a five-frame history of
proprioception and command signals, together with a per-EE block containing the commanded pose,
compliant reference, measured force, and impedance parameters, under biased and noisy sensor models.

\begin{figure*}[!t]
\centering
\includegraphics[width=0.99\textwidth]{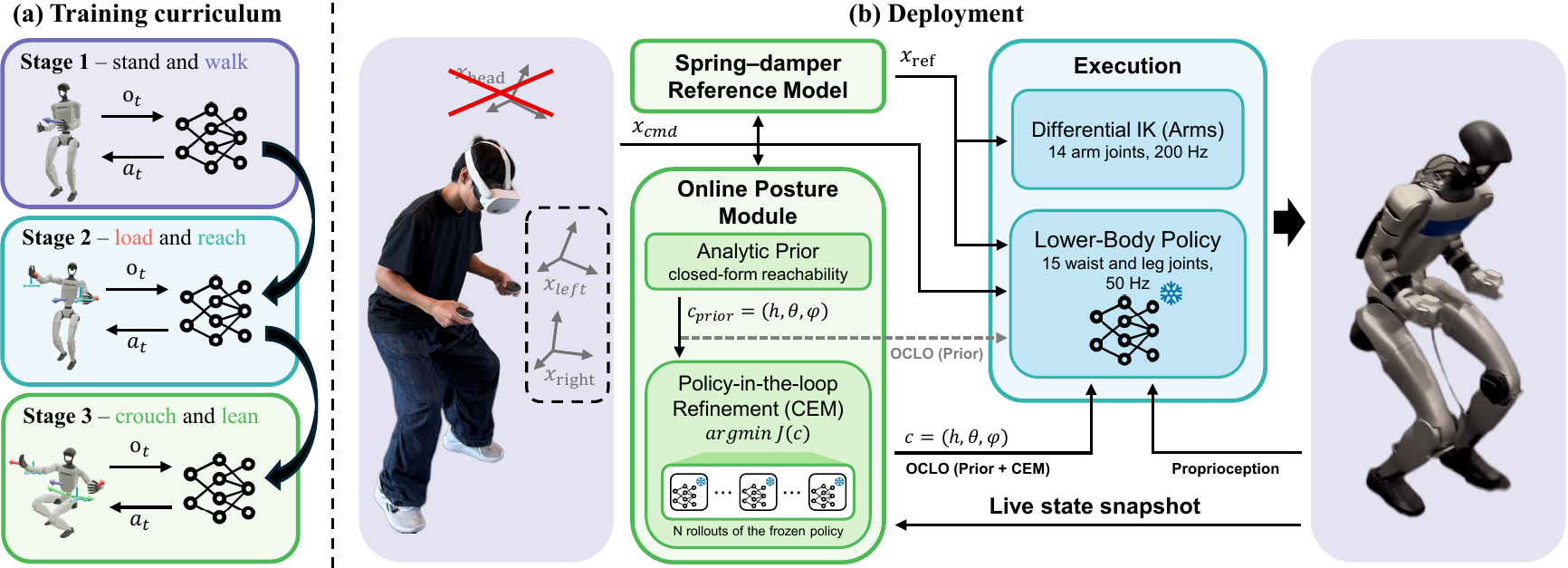}
\caption{\textbf{Training and deployment of \method{}.}
(a) The lower-body policy is trained through a three-stage curriculum:
locomotion, EE reaching with interaction loads, and finally posture adaptation
through crouching and leaning. All training commands are procedurally generated, and the policy is frozen after training. (b) At deployment, two EE targets define the manipulation intent. The spring-damper model produces compliant EE references for differential IK, while the online posture module generates pelvis height and torso orientation. The analytic prior can be applied directly (\method{} (Prior)). When higher pose accuracy is desired, it can be refined by a Cross-Entropy Method (CEM) optimizer using frozen-policy rollouts from a snapshot of the live robot state (\method{} (Prior + CEM)). The resulting posture command is executed by the lower-body policy together with the IK-controlled arms.}
\label{fig:pipeline}
\end{figure*}

At deployment, the operator provides two EE targets through the hand controllers
and a walking velocity through the G1's joystick (Fig.~\ref{fig:pipeline}); no explicit posture command is required. The pelvis height and torso
orientation that three-point interfaces obtain from head or chest tracking~\cite{omnih2o,opentv,mobiletv}
are instead generated online from the two EE targets (Sec.~\ref{sec:posture}).

\subsection{Interface and Compliance}
\label{sec:interface}

All EE commands, references, and forces are expressed in a gravity-aligned frame
$\mathcal{T}$. 
$\mathcal{T}$ moves with the robot, is yaw-aligned with its heading, and has its vertical origin at the support surface. This makes posture adaptation meaningful:
if EE targets were defined relative to the pelvis, lowering the pelvis would also lower
the targets, and crouching would not reduce EE error. Any source providing a pair of
EE poses in $\mathcal{T}$, optionally with an interaction force~\cite{unifiedforce},
can drive the system.

To make the hands yield under external forces, each EE tracks a compliant reference
$x_{\mathrm{ref}}$ rather than the commanded pose $x_{\mathrm{cmd}}$. The reference
is generated by a virtual mass-spring-damper,
\begin{equation}
\resizebox{0.9\columnwidth}{!}{$
m_v\,\ddot{x}_{\mathrm{ref}}
=
k_p\,(x_{\mathrm{cmd}}-x_{\mathrm{ref}})
+
k_d\,(\dot{x}_{\mathrm{rest}}-\dot{x}_{\mathrm{ref}})
+
\mathrm{sat}_a(F_{\mathrm{ext}})
$}
\label{eq:spring}
\end{equation}
with an analogous rotational system. Here, $m_v$ controls the virtual inertia,
$k_p$ pulls the reference toward the commanded pose, and $k_d$ damps its motion
relative to the base-following rest velocity $\dot{x}_{\mathrm{rest}}$.
The external force $F_{\mathrm{ext}}$ displaces the reference through
$\mathrm{sat}_a$, which limits large out-of-distribution forces while preserving
their direction ($a=\SI{40}{\newton}$). Stiffness is randomized per hold,
$k_p\sim\mathcal{U}[12,24]$, with near-critical damping, and the impedance
parameters are included in the policy observation so that the policy can adapt to
the current level of compliance~\cite{facet,softmimic}.

The arms track the compliant references using damped least-squares differential
inverse kinematics~\cite{nakamura}, with soft joint-limit and posture regularization
together with gravity and load feed-forward. Changes in pelvis height are
automatically accommodated by recomputing the arm configuration.

\subsection{Procedural Commands and Reward}
\label{sec:commands}

All training commands are procedurally generated rather than replayed from a motion
corpus. Base velocity is resampled every \numrange{3}{8}\,\si{\second}, with episodes
split among standing, turning, and walking and ranges widened by a performance
gate~\cite{leggedgym}. Posture commands
$c_{\mathrm{post}}=(h,\theta,\varphi)$ specify pelvis height, torso pitch, and torso
roll and are resampled on the same schedule, with a speed-dependent lower bound on
pelvis height. EE targets are generated in $\mathcal{T}$ under two anchorings.
Some are anchored to the support surface at an absolute height, forcing the lower
body to crouch for low targets, while the remainder are anchored to the commanded
posture to avoid conflict between EE and posture commands. Loads are applied through
Eq.~\ref{eq:spring} as sustained forces and short impulses whose magnitude increases
with the curriculum~\cite{falcon}.

The reward (Table~\ref{tab:rewards}) combines bounded tracking terms
$r_i=\exp(-e_i^2/\sigma_i^2)$, gait shaping, and standard regularization
~\cite{leggedgym}. Three design choices are central to our system.
\emph{First, EE tracking is evaluated against the compliant reference.}
Because the policy does not control the arms, it can reduce EE error only through
lower-body and waist motion, encouraging body shifts and crouching to extend reach
and accommodate external loads~\cite{deepwbc}.
\emph{Second, posture tracking replaces fixed-stance penalties.}
In the final stage, fixed base-height and flat-orientation penalties are replaced by
tracking of the commanded pelvis height and torso orientation~\cite{hugwbc,amo}.
\emph{Third, the gait is shaped rather than imitated.}
Phase-consistency, single-support-duration, and swing-height terms shape a
flat-footed gait without a reference motion trajectory~\cite{exbody,gmt}.

\begin{table}[t]
  \centering
  \caption{\textbf{Principal reward terms.} Tracking terms use $r_i=\exp(-e_i^2/\sigma_i^2)$,
  and Stage~3 swaps the fixed-stance penalties (highlighted rows replace them) for
  posture tracking. Standard regularization terms (action rate, joint acceleration, velocity,
  torque, limits, contacts) are omitted.}
  \label{tab:rewards}
  \footnotesize
  \setlength{\tabcolsep}{4pt}
  \renewcommand{\arraystretch}{1.15}
  \resizebox{\columnwidth}{!}{%
  \begin{tabular}{@{}llccc@{}}
  \toprule
  \rowcolor{headhl}
  \textbf{Term} & \textbf{Error $e_i$} & $\sigma_i$ & $w_i$ & \textbf{Stage}\\
  \midrule
  \multicolumn{5}{@{}l}{\textit{End-effector tracking}}\\
  \quad EE position (per arm)    & $\norm{p_{ee}-p_{\mathrm{ref}}}$ & $0.2$\,m & $+0.5$ &
  2--3\\
  \quad EE orientation (per arm) & quat.\ geodesic                  & $0.5$    & $+0.5$ &
  2--3\\
  \addlinespace[3pt]
  \multicolumn{5}{@{}l}{\textit{Locomotion}}\\
  \quad Velocity command (lin.\,/\,yaw) &
  $\norm{v^{xy}_b-\vc^{xy}}$\,/\,$|\omega^z_b-\omega^z_{\mathrm{cmd}}|$ & $0.35$\,/\,$0.25$ &
  $+4.0$\,/\,$+2.5$ & 1--3\\
  \addlinespace[3pt]
  \multicolumn{5}{@{}l}{\textit{Posture \& base}}\\
  \quad Base height\,/\,flat orient.\ (fixed) &
  $(z_{\mathrm{pelvis}}-0.78)^2$\,/\,$\norm{g^{xy}_b}^2$ & --- & $-10.0$\,/\,$-5.0$ & 2\\
  \quad Pelvis height tracking    & $|h-h_{\mathrm{cmd}}|$      & $0.10\!\to\!0.05$ &
   $\mathbf{+2.5}$ & 3\\
  \quad Torso orientation tracking & $\norm{g_b-g^{*}_b}$        & $0.2$             &
   $\mathbf{+1.5}$ & 3\\
  \quad Knee--height consistency           & knee flexion vs.\ $h_{\mathrm{cmd}}$ & ---
  & $-2.0$          & 3\\
  \addlinespace[3pt]
  \multicolumn{5}{@{}l}{\textit{Gait \& feet}}\\
  \quad Contact phase consistency & clock vs.\ contact state & --- & $+2.0$ & 1--3\\
  \quad Single-support duration   & tent peaked at $0.25$\,s & --- & $+2.0$ & 1--3\\
  \quad Swing ht.\,/\,toe att.\,/\,slide & $(z_{\mathrm{sole}}-0.03)^2$\,/\,clearance\,/\,slip
  & --- & $-40$\,/\,$-5$\,/\,$-2$ & 1--3\\
  \bottomrule
  \end{tabular}%
  }
  \end{table}

\subsection{Three-Stage Curriculum}
\label{sec:training}

We train with PPO~\cite{ppo} in GPU-parallel MuJoCo Warp environments
~\cite{mjwarp,mjlab}, with domain randomization over friction, link mass and
center of mass, actuator delay and gains, wrist payload, and sensor bias~\cite{tobin}.
Each stage warm-starts from the previous stage, and progression is gated on sustained
performance (Fig.~\ref{fig:pipeline}a).

\textbf{Stage 1} learns standing, walking, and turning from velocity commands with
the arms parked.
\textbf{Stage 2} adds EE targets and interaction loads while keeping the posture near nominal, allowing the policy to learn lower-body support for reaching and compliant interaction.
\textbf{Stage 3} replaces fixed-stance penalties with posture tracking and expands
the posture range toward deep crouching and forward leaning. Low EE targets are
introduced together with lower pelvis heights so that crouching and low reaching are
learned jointly.

\subsection{Online Posture Generation}
\label{sec:posture}
\vspace{-1mm}

 Posture commands are sampled
independently of EE targets during training, allowing the trained policy to accept
posture commands generated online at deployment, either by the analytic prior or by a Cross-Entropy Method (CEM) optimizer~\cite{cem}. We evaluate three variants:
\method{} (Prior), \method{} (CEM), and \method{} (Prior + CEM).

\textbf{Analytic prior.}
We first define a closed-form reachability prior,
\begin{equation}
h=\max\!\left(h_0-k_h\,\delta,\;h_{\min}\right),\
\theta=k_\theta\,\delta+k_\eta\,\eta,\
\varphi=0,
\label{eq:prior}
\end{equation}
where $\delta=\relu{z_0-\min_j z_j}$ measures how far the lower EE target lies
below standing reach, and $\eta=\relu{\max_j x_j-x_0}$ measures the forward excess
of the farther target. The prior lowers the pelvis for low targets and increases torso pitch for low or forward-reaching targets. The lower hand determines $\delta$, so the pelvis response is driven by the target requiring the deeper crouch. Because the prior does not model balance, $h_{\min}$
prevents excessively low pelvis commands. Its constants are fitted once from a
free-torso reachability sweep and then fixed for all tasks. The prior runs in
microseconds on a CPU.

\textbf{Policy-in-the-loop refinement.}
The analytic prior considers reachability but does not evaluate the closed-loop
consequences of a posture. We therefore refine it by evaluating candidate posture
commands with the frozen policy in the loop~\cite{mjpc,sumo,rgbmppi}.
Each candidate is a constant posture command $c$ held over the rollout horizon.
The current deployed state, including the compliance integrators, is copied into simulations that run in parallel CPU threads as in Sumo~\cite{sumo}. These rollouts reproduce the plans and closed-loop success of the GPU-batched version~\cite{mjwarp} used for Table~\ref{tab:simulation} (Sec.~\ref{sec:posture_source}). Each rollout is evaluated by

\vspace{-3mm}
\begin{equation}
\begin{aligned}
J(c)={}& w_f\,\mathbbm{1}[\mathrm{fall}]
+\frac{1}{T}\sum_{t=1}^{T}
\Big(
w_e C^{ee}_t +
w_c C^{cp}_t +
w_p C^{\mathrm{post}}_t \\
&\qquad +
w_v C^{\mathrm{vel}}_t +
w_\tau C^\tau_t +
w_l C^{\mathrm{lim}}_t
\Big) \\
&+
w_r\norm{\Lambda(c-c_{\mathrm{prev}})}^2
+
w_s\norm{\Lambda(c-c_{\mathrm{prior}})}^2 .
\end{aligned}
\label{eq:cost}
\end{equation}

Here, $T$ is the number of valid rollout steps up to the planning horizon
or termination by a fall, and $\Lambda$ is a diagonal scaling matrix that places
pelvis height and torso angles on comparable scales.

The cost captures four factors. First, the fall penalty and capture-point cost
$C^{cp}_t$, defined as the distance of the capture point~\cite{capturepoint}
outside the support region, penalize unstable postures. Second, EE error
$C^{ee}_t$ penalizes candidate postures from which the compliant EE references
cannot be accurately tracked, while $C^{\mathrm{post}}_t$ penalizes posture
commands that the controller cannot realize. Third, velocity error, lower-body
torque, and joint-limit violations penalize postures that interfere with
locomotion or require excessive effort. Finally, regularization toward the
previously deployed posture $c_{\mathrm{prev}}$ suppresses chattering, while
regularization toward the analytic prior $c_{\mathrm{prior}}$ keeps the search
near the prior solution.

All weights are fixed across tasks, and the cost contains no object- or
task-specific objective term. A receding-horizon cross-entropy
method~\cite{cem} performs one elite update per replan. The analytic prior
initializes the sampling distribution, remains as a deterministic candidate,
and anchors the search through the prior regularizer. Refinement can therefore
adjust the prior using closed-loop stability and control information, including
torso roll, which the analytic prior fixes to zero.

\subsection{Two-Point Interface and Deployment}
\label{sec:deploy}

The operator uses two hand controllers, providing only their poses and one clutch
input per hand, and the G1's joystick sets the walking velocity; no head, chest, or hip tracking and no exoskeleton are required
~\cite{omnih2o,opentv,homie}. While a clutch is held, each EE command follows the
controller displacement relative to the pose at clutch engagement. This relative
mapping avoids calibration between the operator's stance and the robot.

On hardware, base velocity is estimated kinematically from the planted foot,
impedance parameters are fixed at the center of the trained range, and interaction
forces are estimated from motor torques because the robot has no wrist force sensors. In simulation, showing the policy this center value or the mirrored value $36-k_p$ instead of the true stiffness changed success by at most 3.3 points and EE error by under 1\% (EE Push and Sustained Load, 90 paired episodes each).
Force estimates are gated near joint limits, where contact with a joint stop can be
difficult to distinguish from an external force. In simulation the palm roughly follows the commanded stiffness when the force is known, but the torque-based estimate in the loop makes it stiffer and noisier, and loads near \SI{1}{\newton} read zero.


\section{Experimental Evaluation}
\label{sec:experiments}

Three questions organize the evaluation. \textbf{Q1} examines whether whole-body posture must be generated online and how to generate it from two EE targets using the analytic prior, policy-in-the-loop sampling, or both, and whether the module transfers to a human-motion-pretrained controller. \textbf{Q2} identifies which procedurally generated training signals are necessary and how their removal affects behavior. \textbf{Q3} evaluates hardware disturbance rejection and loco-manipulation beyond a fixed stance. Simulation addresses Q1 and Q2 with paired, repeatable episodes, while hardware addresses Q3 and tests whether the Q2 differences persist on the robot.

\subsection{Simulation Protocol}
\label{sec:sim_protocol}

We evaluate a Unitree G1 in mjlab~\cite{mjlab}, with the policy running at \SI{50}{\hertz} and inverse kinematics at \SI{200}{\hertz}. All episode-level variables—including EE targets, base-velocity commands, and disturbance timing and magnitude—are generated by a sampler keyed only by the pair (seed, episode), ensuring that all methods are evaluated on identical episodes. Each table entry aggregates 90 paired episodes (3 seeds $\times$ 30 episodes), each lasting \SI{12}{\second} after a \SI{1}{\second} warm-up. When posture refinement is enabled, the sampler evaluates $N=64$ candidates over a horizon of $H=\SI{1.0}{\second}$ and is invoked at a nominal rate of \SI{2}{\hertz}. An episode is successful if the robot remains upright, acquires the EE reference within \SI{4}{\second}, and maintains the tracking error below threshold thereafter.

Five tasks evaluate tracking and disturbance rejection across different posture regimes. \textbf{Track} follows a base-velocity command while maintaining two EE references. \textbf{Base Push} and \textbf{EE Push} apply impulsive disturbances to the torso and end-effectors, respectively. \textbf{Sustained Load} applies a continuous load to one hand throughout the episode. \textbf{Crouched Push} applies torso impulses while the robot tracks low EE references and is evaluated by fall rate alone. Table~\ref{tab:simulation} reports task success and EE position and orientation error relative to the compliance-displaced reference.

\textbf{Methods.} We distinguish methods by the source of their posture command. \textbf{\method{} (Prior)} directly applies the analytic reachability posture of Eq.~\ref{eq:prior}. \textbf{\method{} (CEM)} uses sampling without the analytic prior. \textbf{\method{} (Prior + CEM)} is the full posture module, in which the prior initializes, anchors, and re-seeds the sampler. \textbf{\method{} (Nominal)} holds the same policy at a fixed nominal posture, and \textbf{End-to-end} is a policy trained on EE targets alone (Sec.~\ref{sec:posture_source}). \textbf{B1} removes both EE-target sampling and EE-force perturbations during training, while \textbf{B2} retains target sampling but removes force perturbations. Both ablations retain the same posture interface and use the full Prior + CEM module at test time, isolating the effect of the training signals. Finally, \textbf{SONIC (Prior)} and \textbf{SONIC (Prior + CEM)} apply the same posture-generation modules to a pretrained SONIC controller~\cite{sonic}, and \textbf{SONIC (Nominal)} holds it at its nominal height.

\begin{table*}[t]
  \centering
  \caption{\textbf{Simulation evaluation.}
  Success (\%), mean EE position error to the compliant reference (cm), and EE orientation error ($^\circ$) over 90 paired episodes per cell, with
  identical episode draws across methods. The upper block compares training ablations B1 and
  B2 and the End-to-end policy with \method{} under four posture variants; bold indicates the best result
  within this block. The lower block evaluates pretrained SONIC~\cite{sonic} using the same
  posture interface. $^\dagger$For Crouched Push, success is determined by falls only.}
  \label{tab:simulation}
  \footnotesize
  \setlength{\tabcolsep}{3pt}
  \renewcommand{\arraystretch}{1.2}
  \resizebox{\textwidth}{!}{%
  \begin{tabular}{@{}lrrrrrrrrrrrrrrr@{}}
  \toprule
  \rowcolor{headhl}
  & \multicolumn{3}{c}{\textbf{Track}} & \multicolumn{3}{c}{\textbf{Base Push}} &
  \multicolumn{3}{c}{\textbf{EE Push}} & \multicolumn{3}{c}{\textbf{Sustained Load}} &
  \multicolumn{3}{c}{\textbf{Crouched Push}$^{\dagger}$}\\
  \cmidrule(lr){2-4}\cmidrule(lr){5-7}\cmidrule(lr){8-10}\cmidrule(lr){11-13}\cmidrule(lr){14-16}
  \rowcolor{headhl}
  \textbf{Method} & Succ.~$\uparrow$ & Pos.~$\downarrow$ & Ori.~$\downarrow$ & Succ.~$\uparrow$ & Pos.~$\downarrow$ & Ori.~$\downarrow$ & Succ.~$\uparrow$ & Pos.~$\downarrow$ & Ori.~$\downarrow$ & Succ.~$\uparrow$ & Pos.~$\downarrow$ & Ori.~$\downarrow$ & Succ.~$\uparrow$ & Pos.~$\downarrow$ & Ori.~$\downarrow$\\
  \midrule
  B1, no targets, no forces & 56.7 & 9.1 & 14.7 & 13.3 & 30.5 & 43.0 & 32.2 & 12.0 & 17.4 & 41.1 & 11.7 & 16.6 & 4.4 & 46.5 & 68.4\\
  B2, targets, no forces & 56.7 & 4.0 & 23.0 & 23.3 & 26.3 & 45.9 & 40.0 & 6.3 & 24.5 & 70.0 & \textbf{5.7} & 25.4 & 16.7 & \textbf{37.0} & 72.4\\
  End-to-end, no posture command & 43.3 & 6.6 & 21.6 & 15.6 & 18.8 & 26.2 & 37.8 & 8.1 & 21.3 & 54.4 & 8.6 & 21.9 & 45.6 & 44.8 & 55.8\\
  \textbf{\method{} (Nominal)} & 15.6 & 17.2 & 15.3 & 7.8 & 37.4 & 24.3 & 8.9 & 18.9 & 16.0 & 8.9 & 20.1 & 15.0 & 28.9 & 74.1 & 63.7\\
  \textbf{\method{} (CEM)} & 37.8 & 7.0 & \textbf{10.3} & 27.8 & 22.8 & \textbf{17.1} & 25.6 & 8.7 & \textbf{11.6} & 42.2 & 9.1 & \textbf{10.7} & 40.0 & 47.7 & 52.1\\
  \textbf{\method{} (Prior)} & \textbf{82.2} & \textbf{3.7} & 17.6 & 61.1 & 17.7 & 22.6 & 62.2 & \textbf{5.7} & 17.9 & 64.4 & 6.2 & 16.6 & 45.6 & 39.0 & 51.3\\
  \textbf{\method{} (Prior + CEM)} & 77.8 & 4.2 & 11.5 & \textbf{62.2} & \textbf{16.2} & \textbf{17.1} & \textbf{63.3} & 5.9 & 12.5 & \textbf{72.2} & 6.3 & 11.7 & \textbf{48.9} & 37.2 & \textbf{48.2}\\
  \midrule
  SONIC~\cite{sonic} (Nominal) & 30.0 & 12.7 & 35.5 & 22.2 & 22.8 & 40.4 & 23.3 & 13.5 & 36.8 & 38.9 & 14.4 & 37.9 & 45.6 & 58.1 & 62.2\\
  SONIC~\cite{sonic} (Prior) & 81.1 & 6.0 & 32.8 & 63.3 & 12.1 & 37.9 & 71.1 & 7.0 & 34.8 & 90.0 & 7.1 & 36.0 & 68.9 & 29.8 & 52.2\\
  SONIC~\cite{sonic} (Prior + CEM) & 80.0 & 6.0 & 32.8 & 66.7 & 11.9 & 37.0 & 73.3 & 6.5 & 33.8 & 91.1 & 7.1 & 36.1 & 72.2 & 31.2 & 50.4\\
  \bottomrule
  \end{tabular}%
  }
  \end{table*}

\subsection{Producing the Posture Online (Q1)}
\label{sec:posture_source}

\textbf{Posture generation is needed.} At a fixed nominal posture the same policy succeeds less often on every task (Table~\ref{tab:simulation}). On Track it succeeds in 15.6\% of episodes against 82.2\% with the prior. On Track, Base Push, and EE Push it never reaches the targets in 75 of 90 episodes. Its orientation error is lower than the prior's on three tasks even though its palms stay 17 to \SI{20}{\centi\meter} from their references. Low orientation error alone does not mean a good posture.

\textbf{An end-to-end policy learns to crouch.} We trained a policy on EE targets alone from the end of the second training stage, with its posture input fixed at the nominal stance and no posture reward (End-to-end in Table~\ref{tab:simulation}). Its pelvis drops to about \SI{0.55}{\meter} on the standing tasks and \SI{0.36}{\meter} on Crouched Push. It succeeds less often than \method{} (Prior) on four tasks and as often on Crouched Push, and its EE errors are higher on every task.

\textbf{The prior makes sampling usable.} Among the online posture sources, \method{} (CEM) is the weakest on every reach-scored task, achieving 37.8\% success on Track compared with 82.2\% for \method{} (Prior). Its failures are primarily failures to acquire the EE reference within the allowed time rather than falls. Starting from the nominal stance, the short-horizon sampler can only adjust the posture gradually and often cannot reach substantially different configurations, such as a deep crouch, within the few replans available before the acquisition deadline (Sec.~\ref{sec:posture}). The analytic prior instead initializes the search near a posture from which the commanded EE targets are reachable, allowing subsequent sampling to focus on local refinement.

\textbf{Refinement improves pose quality with only modest changes in task success.}
Starting from the analytic prior, sampling reduces EE orientation error on all five tasks, from $17.6^\circ$ to $11.5^\circ$ on Track and from $22.6^\circ$ to $17.1^\circ$ on Base Push. The geometric prior primarily identifies postures from which the commanded EE positions are reachable, but it does not distinguish how well the commanded EE orientations can be realized from different reachable postures. Policy-in-the-loop refinement exposes this difference by evaluating candidate postures through closed-loop rollouts against the compliant reference. Across the four disturbed tasks, refinement changes success by $+1.1$, $+1.1$, $+7.8$, and $+3.3$ percentage points, while Track decreases by $4.4$ points, where the prior alone already achieves high reach success. On the unperturbed tasks, EE position error remains within \SI{0.5}{\centi\meter} of the prior, indicating that the orientation improvement does not come from sacrificing position accuracy. Removing interaction-force information from the rollouts changes success by at most 4.4 points, suggesting that the benefit of refinement comes primarily from evaluating candidate postures in closed loop rather than from explicit force awareness.

\textbf{The module transfers.}
On SONIC~\cite{sonic}, the prior raises success by 23.3 to 51.1 points over its nominal height, and adding refinement to the prior increases success on four of five tasks by 1.1--3.4 percentage points and reduces EE orientation error on three, while changing it by at most $0.1^\circ$ on the other two. SONIC (Prior + CEM) achieves higher task success than \method{} (Prior + CEM) on every task, but its EE orientation error is approximately 2--3$\times$ larger on the four standing tasks. \method{} (Prior) also tracks the compliant reference more closely than SONIC (Prior) on Track, EE Push, and Sustained Load and less closely under torso pushes. Because task success is primarily determined by balance and EE position tracking, it does not fully capture pose quality. We therefore report EE position and orientation error alongside success and use SONIC to evaluate whether the posture-generation module transfers to a separately pretrained controller.

\textbf{Planner budget.}
We select the refinement budget using Crouched Push. Success is 43.3, 45.6, \textbf{48.9}, and 45.6\% for rollout horizons of 0.2, 0.5, 1.0, and \SI{2.0}{\second}; 46.7, \textbf{48.9}, and 45.6\% for 16, 64, and 128 candidates; and 43.3, \textbf{48.9}, and 46.7\% for replanning rates of 1, 2, and \SI{4}{\hertz}.  

Refinement costs more than the prior but is fast enough for online use. Our GPU-batched implementation, used for the experiments above, requires \SI{1.4}{\second} per replan on an RTX A6000; an equivalent CPU implementation that runs the rollouts in parallel threads, as in Sumo~\cite{sumo}, reproduces the GPU implementation's plans and closed-loop success at \SI{68}{\milli\second} per replan on an Intel i9-14900K (\SI{15}{\hertz}), or \SI{42}{\milli\second} with 32 candidates. On the robot this CPU version shares the off-board laptop that runs the controller (Intel Core Ultra 9 185H). With 64 candidates over \SI{1}{\second} it replans in \SI{236}{\milli\second} and keeps up with the \SI{2}{\hertz} rate, and the \SI{50}{\hertz} control loop runs as it does with the prior alone. It picks the same posture as our Python rollouts in 10 of 10 recorded robot states. The Prior + CEM runs of Table~\ref{tab:robustness} used these slower Python rollouts. The analytic prior requires only \SI{0.07}{\milli\second} on a CPU. 

\begin{figure}[t]
\centering
\includegraphics[width=0.95\columnwidth]{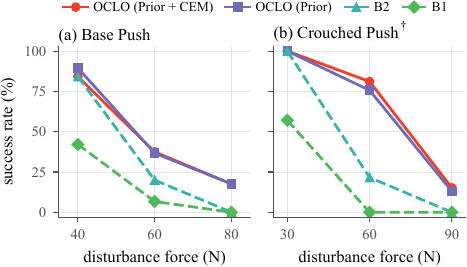}
\caption{\textbf{Success rate under increasing disturbance force.} (a) Base Push at 40, 60, and 80\,N and (b) Crouched Push at 30, 60, and 90\,N. $^\dagger$Crouched Push is scored on falls alone.}
\label{fig:force_ladder}
\end{figure}

\subsection{Which Training Signals the Interface Needs (Q2)}
\label{sec:training_signal}

The success ordering B1 $\le$ B2 $<$ \method{} holds across all tasks in Table~\ref{tab:simulation}, with different failure modes reflecting the training signals removed from each ablation. B1, which is trained without EE-target or EE-force sampling, fails primarily by losing balance rather than by failing to acquire the reference. On Base Push, it falls in 30\% of episodes, and on Crouched Push, which is evaluated by falls alone, in 96\%, compared with 51\% for \method{} (Prior + CEM). 

B2 restores EE-target sampling but not EE-force sampling. This recovers nearly all of the gap on Sustained Load, with 70.0\% success compared with 72.2\% for \method{}, but much less on Base Push (23.3\% vs.\ 62.2\%) and Crouched Push (16.7\% vs.\ 48.9\%). When both use the prior alone, B2 still trails \method{} by 16.7 points on EE Push and by 25.6 on Crouched Push, and it leads by 4.4 on Sustained Load. These results suggest that target sampling is sufficient for much of the static-load robustness, whereas force-related training is important for maintaining balance under transient disturbances. Fig.~\ref{fig:force_ladder} extends this ordering across disturbance magnitude. B2 keeps pace only at the lowest force. At 60\,N, B2 keeps 20\% success on Base Push and 22\% on Crouched Push while \method{} keeps 37 to 38\% and 76 to 81\%, and at the highest force both ablations fail every episode while \method{} still succeeds in 13 to 18\%. Between the two \method{} configurations, refinement adds 5.4 and 2.2 points on Crouched Push at 60 and 90\,N and is within a few points of the prior on Base Push, consistent with Table~\ref{tab:simulation}. We next test whether the same distinction appears under physical disturbances on the robot.

\subsection{Hardware (Q3)}
\label{sec:realworld}

\begin{figure}[t]
\centering
\includegraphics[width=0.99\columnwidth,pagebox=cropbox]{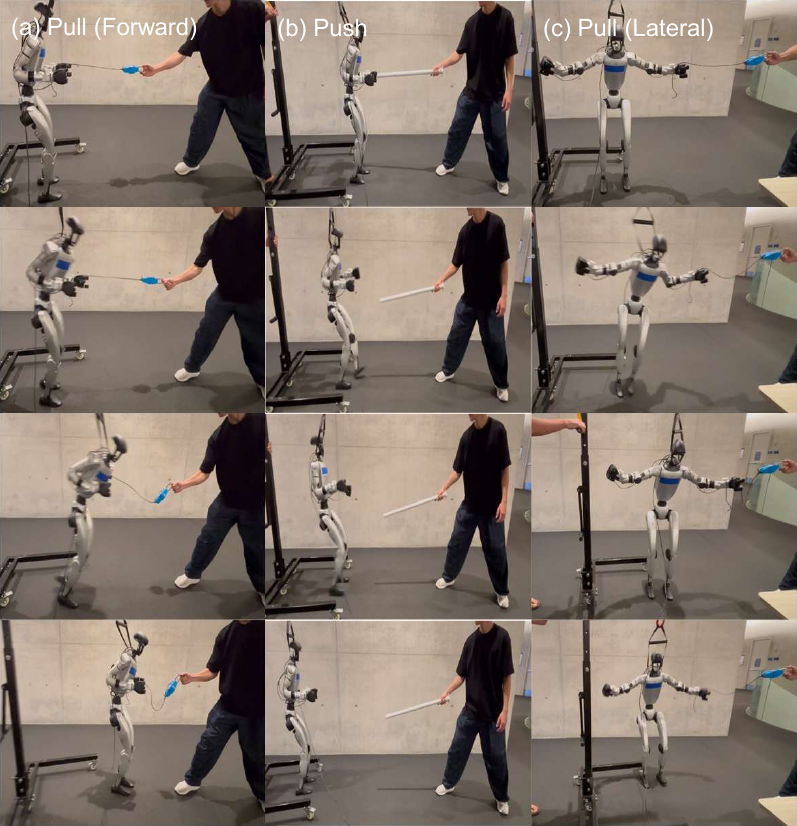}
\caption{\textbf{Evaluation under external force.} The robot maintains balance and recovers its posture under (a) forward pulling, (b) pushing, and (c) lateral pulling from the left. Each column shows representative frames over time for the corresponding disturbance.}
\label{fig:robustness}
\end{figure}

\textbf{Disturbance rejection.}
We compare B1, B2, \method{} (Prior), and \method{} (Prior + CEM) under external disturbances on the robot (Fig.~\ref{fig:robustness}). EE targets remain fixed during each trial. Failure is defined as loss of balance, an unrecoverable step, or intervention by the overhead safety gantry. We apply frontal pull/push disturbances to the torso and frontal pull/push and lateral pull disturbances through the end-effectors.  To rule out weaker loading on \method{} (Prior), we repeated the EE pull at \SIrange{50}{70}{\newton} in addition to the \SIrange{20}{40}{\newton} baseline range; the policy succeeded in 10/10 trials at both levels.

\begin{table}[t]
  \centering
  \caption{\textbf{Real-world disturbance rejection.} Successful trials out of ten per cell.
  Pulls are applied through an inline force gauge at the stated range. Pushes are applied
  manually, the EE push through a PVC pipe, and are not instrumented. B1, B2, and \method{}
  (Prior) hold the same commanded posture throughout, and \method{} (Prior + CEM) refines it
  online. A dash marks an untested condition.}
  \label{tab:robustness}
  \footnotesize
  \setlength{\tabcolsep}{5pt}
  \renewcommand{\arraystretch}{1.2}
  \resizebox{\columnwidth}{!}{%
  \begin{tabular}{@{}lcccccc@{}}
  \toprule
  \rowcolor{headhl}
  \textbf{Method} & \textbf{Torso pull} & \textbf{Torso push} & \multicolumn{2}{c}{\textbf{EE
  pull}} & \textbf{EE push} & \textbf{Lateral pull}\\
  \cmidrule(lr){4-5}
  \rowcolor{headhl}
  & 40--60\,N & & 20--40\,N & 50--70\,N & & 40--60\,N\\
  \midrule
  B1 & 8/10 & 7/10 & 1/10 & --- & 0/10 & 0/10\\
  B2 & 8/10 & 8/10 & 5/10 & --- & 3/10 & 2/10\\
  \textbf{\method{} (Prior)} & \textbf{10/10} & \textbf{9/10} & \textbf{10/10} & \textbf{10/10} &
  \textbf{9/10} & \textbf{8/10}\\
  \textbf{\method{} (Prior + CEM)} & 9/10 & \textbf{9/10} & 9/10 & --- & 8/10 & \textbf{8/10}\\
  \bottomrule
  \end{tabular}%
  }
  \end{table}

All policies remain comparably robust to torso disturbances (Table~\ref{tab:robustness}). The gap is much larger for EE disturbances. B1 and B2 achieve 1/10 and 5/10 on frontal EE pull, 0/10 and 3/10 on EE push, and 0/10 and 2/10 on lateral pull, compared with 10/10, 9/10, and 8/10 for \method{} (Prior). The B2 comparison most directly isolates the effect of EE-force sampling, while B1 also removes EE-target sampling. Given 10 trials per condition, we draw our main hardware claim from the large separation on EE disturbances rather than the small differences on torso disturbances.

Online refinement performs similarly to the prior across the five conditions. Because the prior alone already provides this robustness, we use \method{} (Prior) for the loco-manipulation tasks below.

\textbf{Loco-manipulation tasks.}
Fig.~\ref{fig:teaser} shows \method{} (Prior) on nine qualitative hardware tasks, with full sequences in the supplementary video. Most require online posture adaptation beyond a fixed nominal stance. Together, they span forward, backward, and lateral locomotion, turning, crouching, grasping, insertion, and pushing, pulling, and wiping using two hand controllers, the G1's joystick, and no explicit posture command.

\section{Discussion \& Conclusion}
\label{sec:conclusion}

We presented \method{}, a human-motion-free humanoid loco-manipulation system that separates task-level end-effector intent from whole-body support. The system takes two EE targets as its manipulation interface, generates pelvis height and torso orientation online using an analytic reachability prior with optional policy-in-the-loop refinement, and realizes compliant interaction through force-displaced EE references
together with a lower-body policy trained to accommodate the resulting motion.

The results highlight two complementary strengths of \method{}. First, the posture module automatically selects pelvis height and torso orientation from two EE targets. The analytic prior substantially improves reference acquisition over a fixed nominal posture and over policy-in-the-loop sampling alone, while refinement consistently reduces EE orientation error with only modest changes in task success. The same posture-generation module also improves a separately pretrained SONIC controller, indicating that it is not tied to a particular base policy. Second, force-conditioned training is important for interaction robustness. Ablations trained without EE-target or force variation are substantially less robust, particularly to disturbances applied through the end-effectors. 

Several limitations remain. The posture interface currently controls only pelvis height and torso pitch and roll, excluding stance width, foot placement, gait timing, and online compliance adaptation. Arm redundancy is handled separately by differential inverse kinematics and may degrade near singularities. Policy-in-the-loop refinement relies on simulated rollouts of a robot model and runs on an off-board CPU, and our hardware evaluation of refinement uses fixed targets. In addition, interaction forces are estimated from motor torques rather than measured with wrist force sensors, impedance parameters are fixed at deployment, and the hardware experiments use a single operator. Finally, although the two-point interface can
accept EE poses from arbitrary upstream sources, we evaluate it only through teleoperation. Future work will extend posture generation to foot placement and gait decisions, jointly optimize posture and arm redundancy, adapt compliance online, and integrate the interface with learned upstream policies.

Overall, \method{} demonstrates that humanoid loco-manipulation need not obtain whole-body posture from human motion, an explicit posture command, or a task-specific planner. Instead, task-level EE targets can be complemented by online posture generation, while a common lower-body controller provides the physical support needed to realize them under interaction. This separation allows the same controller to support different postures and upstream commands without retraining.

\end{document}